\documentclass[11pt]{article}

\usepackage[preprint]{acl}

\usepackage{times}
\usepackage{latexsym}

\usepackage[T3, T1]{fontenc}
\usepackage[utf8]{inputenc}

\usepackage{tipa}

\usepackage{microtype}

\usepackage{inconsolata}

\usepackage{graphicx}
\usepackage{booktabs}
\usepackage{amsmath}
\usepackage{multirow}

\usepackage{graphicx}
\usepackage{amsmath}
\usepackage{textgreek}
\usepackage{tikz}

\usetikzlibrary{arrows.meta,shapes.geometric,shapes.symbols,shadows}

\usetikzlibrary{shapes, arrows, positioning, calc}
\usepackage{pgfplots}
\pgfplotsset{compat=1.17}
\usepackage{subcaption} 
\usepackage{microtype}
\usepackage[utf8]{inputenc}
\usepackage{newunicodechar}
\newunicodechar{̂}{\^}

\title{Dealing with the Hard Facts of Low-Resource African NLP}

\author{
 \textbf{Yacouba Diarra},
 \textbf{Nouhoum Souleymane Coulibaly},
 \textbf{Panga Azazia Kamaté},\\ 
 \textbf{Madani Amadou Tall},
  \textbf{Emmanuel Élisé Koné},
  \textbf{Aymane Dembélé},
  \textbf{Michael Leventhal}\\
 RobotsMali AI4D Lab, Bamako, Mali;
  \small{
    \textbf{Correspondence:} \href{mailto:research@robotsmali.org}{research@robotsmali.org}
  }
}
\begin{document}
\maketitle
\begin{abstract}
Creating speech datasets, models, and evaluation frameworks for low-resource languages remains challenging given the lack of a broad base of pertinent experience to draw from. This paper reports on the field collection of 612 hours of spontaneous speech in Bambara, a low-resource West African language; the semi-automated annotation of that dataset with transcriptions; the creation of several monolingual ultra-compact and small models using the dataset; and the automatic and human evaluation of their output. We offer practical suggestions for data collection protocols, annotation, and model design, as well as evidence for the importance of performing human evaluation. In addition to the main dataset, multiple evaluation datasets, models, and code are made publicly available.
\end{abstract}

\section{Introduction}
\label{sec:intro}

End-to-end ASR (E2E-ASR) systems for languages with large amounts of text data, especially English, have achieved human-level performance on several benchmarks \cite{humanparityASR}. In contrast, training E2E-ASR for low-resource languages remains challenging due to the considerable amounts of labeled data and computational resources required by modern deep learning architectures.\cite{kaplan2020scalinglawsneurallanguage}.

Until recently, African language aligned data for ASR existed only as a minute portion of large multilingual datasets, often primarily for benchmarking purposes (\citealp{ardila-etal-2020-common}; \citealp{goyal-etal-2022-flores}), rather than for training models intended to be deployed in systems that recognize real-world speech. No open ASR model exists for the vast majority of the 2000+ languages on the continent.

In pursuit of what they call \textit{omnilingualism}, Meta released the Massively Multilingual Speech \cite{pratap2023scalingspeechtechnology1000} and Omnilingual ASR \cite{omnilingualasr2025} model suites in 2023 and 2025. Their approach used massive self-supervised learning \cite{baevski2020wav2vec20frameworkselfsupervised} and fine-tuning on small labeled datasets, consisting, principally, of publicly available readings of religious texts in the 2023 release, and data obtained from community-centered crowdsourced data in the 2025 release.\footnote{Meta released part of their labeled corpus openly, offering spontaneous speech recordings and their transcriptions for 348 under-served languages, along with training script configuration and docs: https://github.com/facebookresearch/omnilingual-asr/} This project provided some level of ASR capability for many African and non-African languages for the first time. While a positive development, there is less than 50 hours of data for many of those languages, some with less than 10 hours, a very small fraction of the $\mathbf{120,710}$ hours on which the supervised fine-tuning (SFT) models were trained. The underlying self-supervised encoder (a 7B-parameter wav2vec 2.0 model) was trained on approximately 4.3M hours of unlabeled audio \cite{omnilingualasr2025}.

For Bambara, a Manding language spoken in several West African countries (primarily in Mali), with more than 15 million L1 and L2 speakers and mutual intelligibility with Malinke, Dioula, and Mandinka, which are spoken by an additional 25 million people, the development of speech recognition technology could affect a population of roughly 40 million \cite{ethnologue2023}. However, as a \textit{low-literacy, predominantly oral language}, Bambara transcription is a hard problem: few speakers can write it, and even those who can lack the facility to do so quickly and easily \cite{diarra2025costanalysishumancorrectedtranscription}.

The CMU Wilderness Multilingual Speech Dataset, a dataset of aligned sentences and audio for some 700 languages based on readings of the New Testament, is, to the best of our knowledge, the first mention of Bambara in speech corpora prepared specifically to train speech synthesis models \cite{cmuwilderness_black}. The dataset was never released as an open resource. Jeli-ASR, a corpus of 30 hours of griot narrations with their transcriptions and French translations, has so far been the only open ASR dataset for Bambara \cite{Diarra2022Griots}. Since its release in 2022, Jeli-ASR has given rise to derivative datasets and has supported the development of the first openly released ASR models for Bambara on HuggingFace\footnote{\href{https://huggingface.co/oza75}{oza75} released a finetuned Whisper model in early 2024 (later taken down), followed by several releases by \href{https://huggingface.co/RobotsMali/models}{RobotsMali} in early 2025}.

The African Next Voices project (ANV), undertaken by a network of African universities and organizations, recently released what is thought to be the largest dataset of African languages for AI so far \cite{Marivate2025African}, with more still to be published. The project aims to record and transcribe over 9,000 hours of speech in 18 languages across South Africa, Kenya, Nigeria, and Mali (after Bambara was completed the list as a later addition).

In this paper, we present the Bambara portion of this initiative for which we have collected and annotated $612$ hours of spontaneous Bambara speech collected across the southern part of Mali. We share statistics and metadata about the dataset, the collection process and the results of our ASR experiments with models finetuned and tested on a subset of 101 hours.

\section{Data Collection and Annotation}
\label{sec:data-collection}

In the audio recording phase, we followed an approach similar to \citeauthor{emezue2025naijavoicesdatasetcultivatinglargescale}, using \textit{facilitators}: individuals with knowledge of the language—L1 or L2 speakers—whom we trained in data collection guidelines and in the use of our mobile data collection app.\footnote{We also open-source this app, a minimalist Flutter-based tool designed with a simple user interface to minimize user training time: https://github.com/RobotsMali-AI/Africa-Voice-App} The guidelines covered requirements for the recording environment, quality checks for background noise, and management of participants’ contributions with respect to voice quality, staying on topic, and minimizing code-switching to French. Pronunciation, often an issue when recordings are based on read speech, was rarely a concern here, as we recorded spontaneous speech on familiar topics from L1/L2 contributors. In total, we collected 626.32 hours of audio and processed 612 hours to create the dataset.

The raw recordings were then segmented using Silero VAD's open voice activity detection model \cite{SileroVAD}, retaining on average 70\% of the original duration and yielding 423 hours of speech chunks ranging from 240 milliseconds to 30 seconds. This step also removed long silences and inaudible speech from the recordings, increasing the amount of usable speech for the transcription pipeline and eliminating manual segmentation \cite{li2019santlrspeechannotationtoolkit}. Almost all segments are mono-speaker, although a small number of recordings include brief facilitator speech; overlaps are rare. The segments were first pre-transcribed with \href{https://huggingface.co/RobotsMali/soloni-114m-tdt-ctc-v0}{RobotsMali/soloni-114m-tdt-ctc-v0}, and human transcribers were tasked with reviewing and correcting these model-generated transcriptions rather than transcribing from scratch. We then fine-tuned \href{https://huggingface.co/RobotsMali/soloni-114m-tdt-ctc-v2}{RobotsMali/soloni-114m-tdt-ctc-v2} on 98 hours of human-corrected transcripts and re-transcribed the segments with this model. WER and CER metrics were used to compare the two sets of transcriptions, and human review and correction continued on the new model outputs for a period of time to assess their impact on the annotation process (Section~\ref{subsec:human-eval}). Further details on transcription guidelines and the labeling interface are provided in Appendices~\ref{sec:trans-guide} and~\ref{sec:labeling}.

\section{The ANV Bambara Dataset}
\label{sec:dataset}

\begin{figure*}[t]
    \centering
    \includegraphics[width=\linewidth]{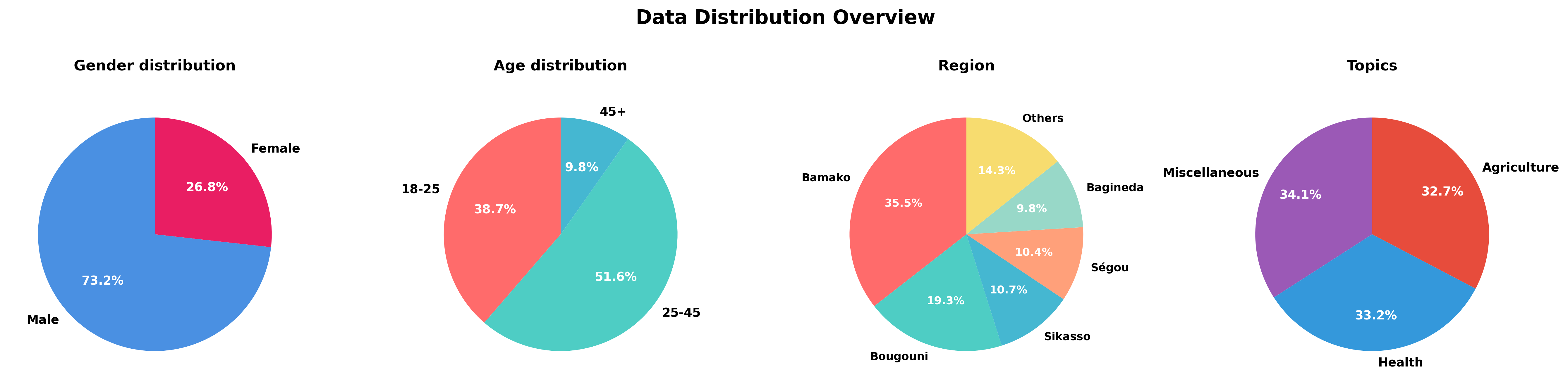}
    \caption{Statistics overview charts of the African Next Voices Bambara dataset: Age, Gender, Region and topics distribution. The first three charts are calculated with respect to the number of speakers while the topics distributions are expressed in durations. The locations represented as 'others' refer rural areas/villages around the 5 other main region}
    \label{fig:fig1}
\end{figure*}

The African Next Voices Dataset is the largest open Bambara ASR corpus collected to date. Comprised of natural, spontaneous speech, often from people with great knowledge in domains such as health, agriculture, the food industry, art, and more, we have captured a profound and authentic snapshot of the Malian society and culture in relatively pure Bambara, with the habitual code-switching to French minimized. The dataset collection was also designed to limit the variance of accents and speaking styles, focusing on the southern regions of Mali, relatively close to the capital. While a truly representative dataset would include code-switching and regional accent variation, the objective fixed for ANV was to gather homogeneous data that would simplify training and provide a baseline for a group of low-resource languages with limited NLP support.

\subsection{Profile of the Dataset}
The dataset features 512 unique speakers from Bamako, the capital, and four localities between 25 and 300 km from Bamako: Ségou, Sikasso, Bagineda, and Bougouni. Figure~\ref{fig:fig1} presents an overview of the gender, age, region, and subject-matter distributions in the dataset. The “Miscellaneous” category encompasses topics ranging from education to social norms and beliefs, history, industry, art, and fashion. The speaker distribution is not gender-balanced, reflecting cultural and security constraints at the time of collection.

The average segment duration is $\approx 2$ seconds. Less than one second chunks were not uploaded in the annotation pipeline as they often consist of formulaic expressions, discourse markers and routine formula of 1 to 3 words such as \textcolor{blue}{nse, nba} (female or male response to a salutation), \textcolor{blue}{ayiwa} (a term used to express agreement or cloture) or \textcolor{blue}{nka} (but). These short segments are transcribed accurately by the models at much lower error rates than longer, novel utterances \cite{tall_2025_17672774}. They are a significant percentage of the total dataset such that including them would skew WER/CER measurements while not contributing to ASR performance. The Huggingface dataset is divided into 3 subsets totaling $\mathbf{874,762}$ utterances totaling to approximately 423 hours. Each subset contains the audio segments and two sets of accompanying transcriptions labeled either \textbf{v1}, created by soloni-v0, trained mainly on jeli-asr, and \textbf{v2} created by soloni-v2, finetuned from soloni-v0 using 98 hours of segments and human-corrected transcriptions collected over the course of the project.

\begin{itemize}
    \item \textbf{The 'human-corrected' subset}: A 159 hour subset ($260,008$ utterances) with human reviewed, corrected and validated transcriptions. This subset is the only one with a 'text' attribute containing a transcription that has gone through human review and correction.
    \item \textbf{The 'model-annotated' subset}: A 212 hour subset ($355,571$ utterances) that has model-generated transcriptions that have not been reviewed by humans. This subset has only \textbf{v1} and \textbf{v2} labels corresponding to the model used to generate transcriptions.
    \item \textbf{The 'short' subset}: A 52 hour subset ($259,183$ utterances) of duration inferior to a second that we have filtered out from the pool of segments to be annotated. Those short utterances are model-annotated and have \textbf{v1} and \textbf{v2} labels.
\end{itemize}

We have also released the original 612 hours dataset comprised of 1777 raw recordings ranging from 8 seconds to 1.48 hours, with all the associated segment timestamps, the anonymized metadata, the SNR quality check results and all the preprocessing code\footnote{We have made the recordings and metadata available through Google Cloud Storage. The link will be found in the GitHub repository holding the code: https://github.com/RobotsMali-AI/afvoices}.

\subsection{Signal-to-Noise Ratio as a proxy to Audio Quality}
\begin{figure}[h!]
    \centering
    \includegraphics[width=\linewidth]{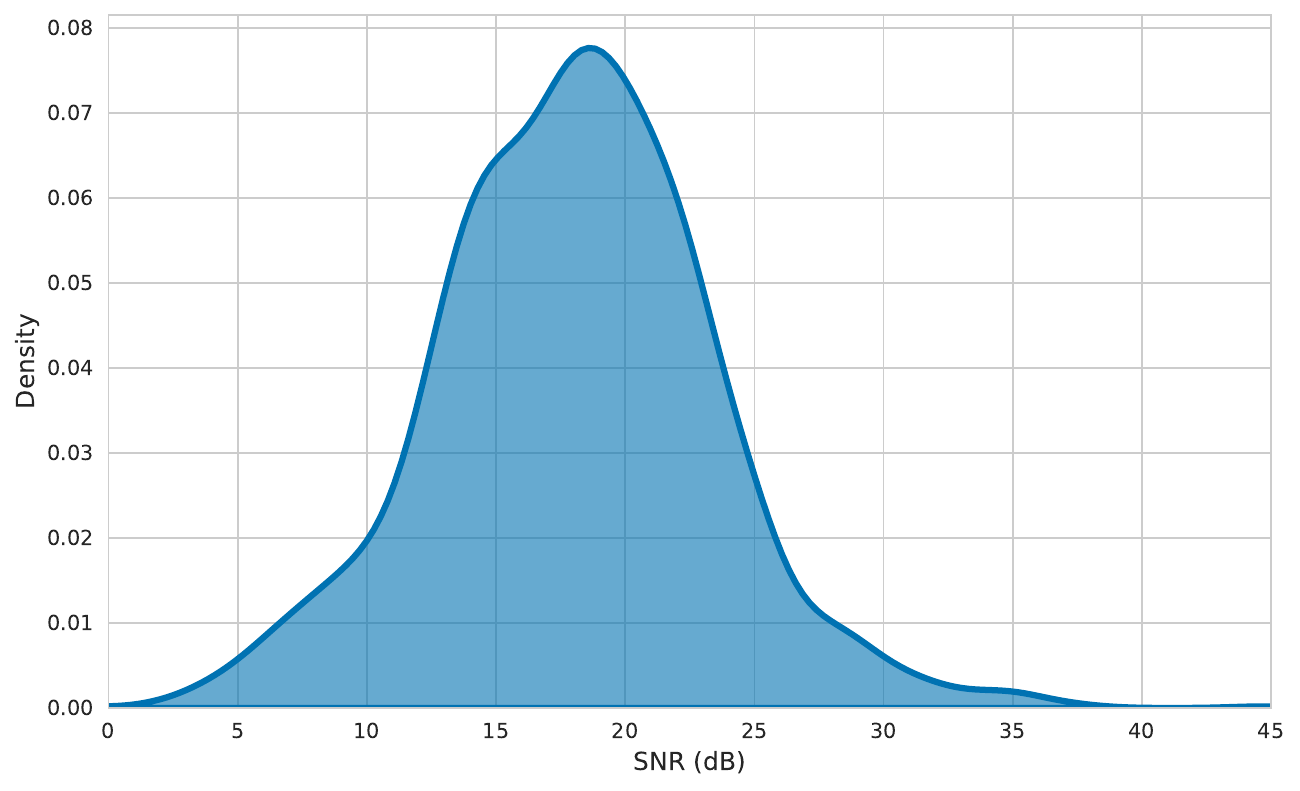}
    \caption{Density Distribution of Signal-to-Noise Ratio values in the African Next Voices Bambara Dataset. Note that the SNR values are not bounded.}
    \label{fig:fig2}
\end{figure}

We use a Voice Activity Detection (VAD) based method to estimate SNR. VAD output is used to separate the signal into two distinct regions: speech activity ($\text{vad}[n]=1$) and voice-inactive ($\text{vad}[n]=0$). We use VAD to estimate the speech and noise power instead of a histogram-based approach such as the standard NIST SNR method. In our setup, SNR is defined, using VAD, as the ratio between the estimated speech power (from speech activity regions) and the estimated noise power (typically the average power in silence regions) from the same recording (\citealp{vad_snr}; \citealp{SileroVAD}.) Figure \ref{fig:fig2} shows the distribution of SNR values from the unsegmented recordings in the dataset.

The classical SNR definition yields a different value distribution than VAD-based estimates, so we follow \citeauthor{xu05b_interspeech}, who report VAD-based SNRs for several noise types and levels at 5, 10, and 20 dB and treat 30 dB as clean speech; Table \ref{tab:snr-distribution} groups our recordings into five bands using thresholds derived from their datapoints. We prefer a VAD-based SNR here because ASR performance depends on the quality of speech frames rather than the entire waveform, including silence. Estimating signal power over VAD-identified speech regions and noise power over non-speech regions yields an SNR measure that better reflects recognition difficulty in our small corpus and allows us to retain and prioritize high-quality speech segments from otherwise noisy files.

\begin{table}[h!]
    \centering
    \begin{tabular}{lcc}
    \toprule
    \textbf{SNR Category} & \textbf{Threshold (dB)} & \textbf{Recordings} \\
    \midrule
    Very low SNR & $< 0$ & 1 \\
    Low SNR & $[0, 5)$ & 15 \\
    Medium SNR & $[5, 15)$ & 486 \\
    High SNR & $[15, 25)$ & 1135 \\
    Very high SNR & $\ge 25$ & 140 \\
    \midrule
    \multicolumn{2}{r}{\textbf{Total Audios}} & \textbf{1777} \\
    \bottomrule
    \end{tabular}
    \caption{Distribution of Audio Recordings by Signal-to-Noise Ratio (SNR) Category.}
    \label{tab:snr-distribution}
\end{table}

71.75\% of the recordings fall into the 'High SNR' to 'very High SNR' categories; this indicates that the dataset consists of relatively clean audio recordings.

\section{ASR Experiments}
\label{sec:asr}
We performed experiments with a subset of our human-corrected transcribed segments to explore the potential of the ANV Bambara dataset for monolingual ASR modeling. We finetuned the models from our earlier experiments with jeli-asr ---themselves finetuned from different models of NVIDIA's Parakeet family of English-trained ASR models and on QuartzNet--- and evaluate all the models on both the test set of our experiment (Afvoices Test) and a smaller, more heterogenous benchmark (Nyana Eval) that we also introduce in this paper. We also report on human evaluation by native speakers, comparing results obtained from the latest finetuned models and their predecessors on the Nyana Eval benchmark (section \ref{subsec:human-eval}).

\paragraph{Experimental setup:} For our experiments, we finetuned open-source models based on NVIDIA's Parakeet family and QuartzNet.  

We finetuned a 114M- and two 600M-parameter Parakeet models\footnote{We trained the Parkeet 600M model with an auto regressive decoder and 600M model with a convolutional decoder. While these models are not further discussed in this paper, all our models with their associated metrics can be found in RobotsMali's HuggingFace repository.}, but as we did not perform human evaluation on the larger model, we only report on the 114M model, \texttt{soloni-114m-tdt-ctc-v0}, in this study. \texttt{soloni-114m-tdt-ctc-v0} uses a Fast-Conformer encoder \cite{rekesh2023fastconformerlinearlyscalable} and a hybrid decoding setup with two independent but jointly trained decoders: a Token-and-Duration Transducer (TDT) decoder ---an extension of the RNN-Transducer that predicts both a token and its duration \cite{xu2023efficientsequencetransductionjointly}---and a convolutional decoder trained with a Connectionist Temporal Classification (CTC) loss \cite{gravesctc2006}. This dual-decoder design makes the model particularly interesting for analyzing how the two decoding approaches behave under the same training conditions. \texttt{soloni-114m-tdt-ctc-v0} also provides insight into how the architecture will perform when scaling up to the larger models in the family.

\texttt{stt-bm-quartznet15x5-v0} is a finetune of NVIDIA's 18M parameter, ultra-compact QuartzNet model, an end-to-end convolutional architecture with 1D time-channel separable convolutions \cite{kriman2019quartznetdeepautomaticspeech}. This model addresses a particularly critical use case in Mali where a large portion of the population does not have access to internet connectivity. We had already deployed this model in a Bambara-language ASR-based reading tutor app\footnote{Our reading tutor app using the QuartzNet model, An B\textipa{\textepsilon} Kalan, is available for \href{https://apps.apple.com/ml/app/an-be-kalan/id6749237637}{iOS} and \href{https://play.google.com/store/apps/details?id=org.robotsmali.literacy_app}{Android}} that runs locally in low-end smartphones.

QuartzNet is a character-based decoding model. The vocabulary for the dataset consists of 38 unique characters, including the 30 letters of the Bambara alphabet, 5 accented French vowels, whitespace, hyphen (used in some compound words) and apostrophe (largely used in writing contractions). For soloni we train a SentencePiece tokenizer with a vocab size of 512. We had 4 NVIDIA A100 GPUs with a combined 320GB of VRAM for the experiment.

\paragraph{Training Data:} We finetuned the two models on 98 hours of voice data with human-corrected transcripts, consisting of $167,816$ utterances, and tested on 3 hours (5175 samples). We implemented and applied most of the normalization steps described by \citeauthor{zupon2021textnormalizationlowresourcelanguages} before training, but we did not remove any of the acoustic event tags (presented in Appendix \ref{sec:trans-guide}) as we wanted to model those events as well.

\paragraph{Training configurations:} We first trained soloni for 110k steps on 2 GPUs, with 32 batch size, using the AdamW optimizer and Noam scheduler with learning rate scaling factor of $0.003$ and a 10\% warmup ratio \cite{vaswani2023attentionneed}. Then we trained the resulting model for 100k more steps on all 4 GPUs, this time with an LR scaling factor of 1.5 and a 2\% warmup ratio, all with no gradient accumulation and \texttt{bf16} precision.

We trained QuartzNet for 65k steps on 4 GPUs, with 64 batch size, using the Novograd optimizer \cite{ginsburg2020stochasticgradientmethodslayerwise} and a Cosine LR scheduler with a $1 \times 10^3$ and $1 \times 10^6$ upper and lower bounds and $6,000$ warmup steps.

\begin{table*}[t]
    \centering
    % Define column format: l (left-aligned) for Model, and 4 x c (centered) for metrics
    \begin{tabular}{l@{\hskip 2cm}cccc} 
        \toprule
        % Header 1: Spanning the two evaluation sets
        \multirow{2}{*}{\textbf{Model}} & \multicolumn{2}{c}{\textbf{WER (\%) $\downarrow$}} & \multicolumn{2}{c}{\textbf{CER (\%) $\downarrow$}} \\
        \cmidrule(lr){2-3} \cmidrule(lr){4-5} % Lines under the evaluation sets
        
        % Header 2: WER/CER under each set
        & \textbf{Afvoices Test} & \textbf{Nyana Eval} & \textbf{Afvoices Test} & \textbf{{Nyana Eval}} \\
        \midrule
        
        % --- soloni-114m (CTC) Group ---
        \multicolumn{5}{l}{\textbf{soloni-114m (CTC)}} \\
        \addlinespace[0.5ex]
        Unfinetuned (v0) & 43.12 & 40.75 & 23.48 & 24.7 \\
        Finetuned (v2) & 29.05 & \textbf{36.07} & 13.41 & \textbf{20.04} \\
        \addlinespace[1ex]

        % --- soloni-114m (TDT) Group ---
        \multicolumn{5}{l}{\textbf{soloni-114m (TDT)}} \\
        \addlinespace[0.5ex]
        Unfinetuned (v0) & 45.52 & 47.1 & 26.68 & 31.27 \\
        Finetuned (v2) & \textbf{28.58} & 38.13 & \textbf{12.94} & 22.3 \\
        \addlinespace[1ex]
        
        % --- Quartznet (CTC) Group ---
        \multicolumn{5}{l}{\textbf{Quartznet (CTC)}} \\
        \addlinespace[0.5ex]
        Unfinetuned (v0) & 73.66 & 65.42 & 37.85 & 30.66 \\
        Finetuned (v2) & 42.57 & 48.97 & 18.70 & 24.22 \\
        \addlinespace[1ex]

        \bottomrule
    \end{tabular}
    \caption{ASR experiment metrics: We apply the same normalization steps to our test sets and this time we remove the acoustic event tags from both the reference and the prediction before calculating the WER and CER. The values in Bold highlight the best performances per metric}
    \label{tab:metrics}
\end{table*}

\section{Evaluatiion of the Models}
\label{sec:results}
We evaluated \texttt{soloni-114m-tdt-ctc-v0} both with the CTC and with TDT decoders, and the QuartzNet model on the Afvoices test set and a smaller benchmark, \textipa{\textltailn \textepsilon na} (transliterated to Nyana for English keyboards and nyana-eval for identification on HuggingFace) \footnote{\textipa{\textltailn \textepsilon na} means “opinion”; this emphasizes the human-evaluation focus of the dataset. We also release it on Hugging Face: https://huggingface.co/datasets/RobotsMali/nyana-eval} , that we compiled through the stratified sampling of 15 audios from each of:
\begin{itemize}
    \item the test set of Kunkado \cite{diarra_kunkado_2025};
    \item a generally unused subset of \textit{jeli-asr} (street interviews) that we cleaned beforehand \cite{Diarra2022Griots};
    \item crowd-sourced recordings of readings of excerpts of the books from the GAIFE project \cite{tapo-etal-2025-gaife}.
\end{itemize}

\subsection{WER evaluation}
\label{subsec:eval}
Table \ref{tab:metrics} presents the Word Error Rates and Character Error Rates of the two models before and after our finetuning experiments. We show a significant improvement for all models across all metrics and benchmarks, up to \textbf{37\% WER improvement} for soloni-114m-tdt, the best performing model overall. The relatively smaller WER improvement on the more challenging Nyana Eval benchmark, containing much noisier---sometimes multi-speaker---audio from street interviews and radio recordings than the Afvoices training and test data, highlights the potential limitations of the model in real-world deployment scenarios.

We note that the CTC branch of soloni lost its edge over TDT branch on the Afvoices test set when we increased the amount of training data from the 30 hours of jeli-asr to 98 hours in this experiment, confirming that, for many sequence modeling tasks, autoregressive architectures outperform non-autoregressive ones when we scale training data (\citealp{graves2012sequencetransductionrecurrentneural}; \citealp{li2020comparisonpopularendtoendmodels}).

\begin{figure}[h!]
    \centering
    \includegraphics[width=\linewidth]{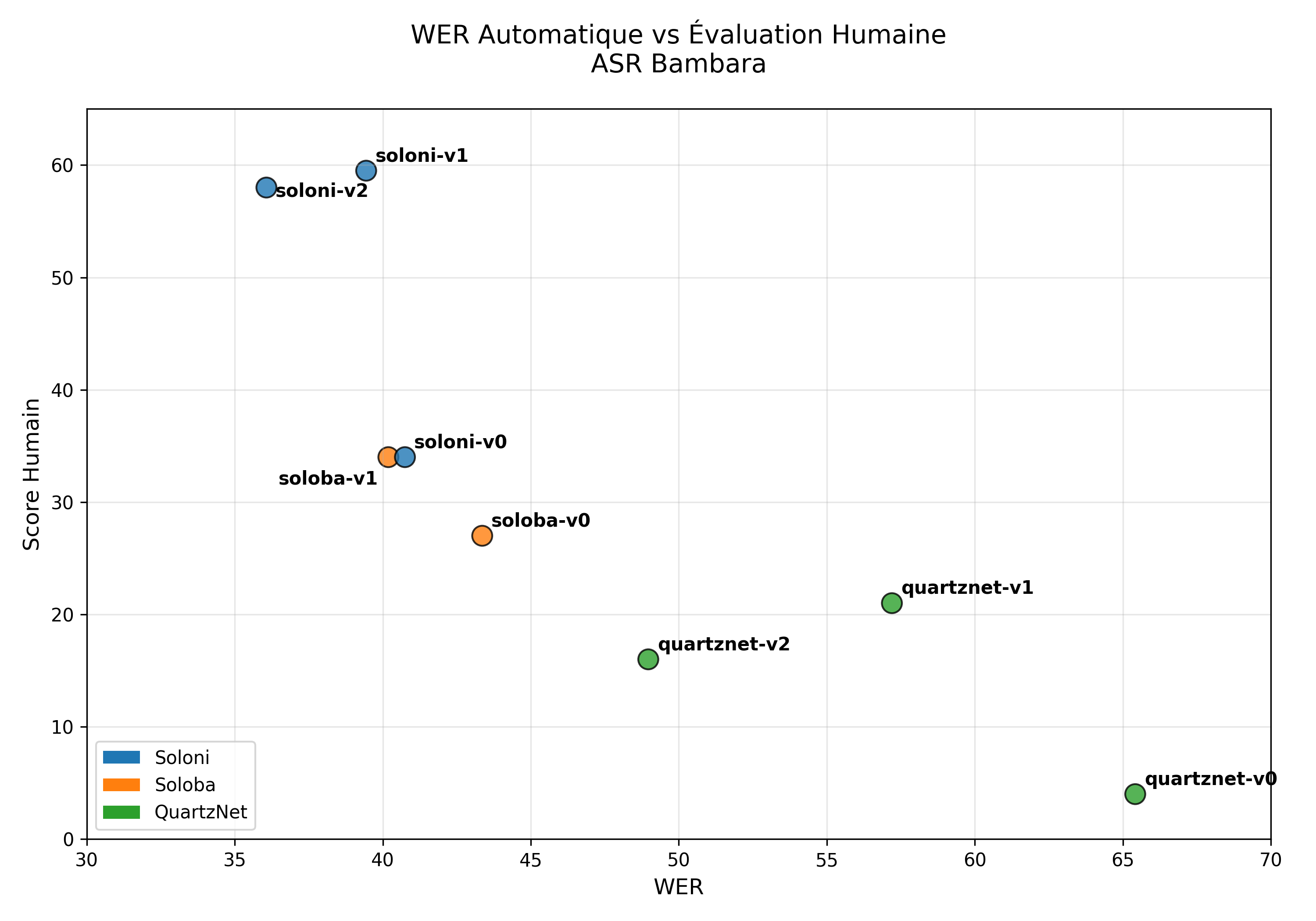}
    \caption{WER vs human evaluation. Figure from \cite{tall_2025_17672774}}
    \label{fig:human-eval}
\end{figure}

\subsection{Human Evaluation}
\label{subsec:human-eval}
We performed a detailed human analysis of the outputs of several RobotsMali ASR models, including the models from this experiment, using Nyana Eval as test data \cite{tall_2025_17672774}. Across models, we observed systematic difficulties with disfluencies, proper names, code switching, and overlapping speech. The highest-rated model in this evaluation is \texttt{soloni-v1}, a finetune of \texttt{soloni-v0} trained on \texttt{RobotsMali/kunkado} \cite{diarra_kunkado_2025}, a dataset composed of everyday speech. \texttt{soloni-v2}, finetuned on the Afvoices dataset with minimized noise, code switching, and voice overlap, was judged slightly less robust by human evaluators on the more natural recordings in Nyana Eval, despite achieving a better WER (36.07\% vs.\ 39.44\%). Figure~\ref{fig:human-eval} plots the WER on Nyana Eval against the corresponding human evaluation scores.

We also report speed gains after replacing the \textbf{v0} transcriptions with \textbf{v2} in the human review-and-correction pipeline. The transcription team completed 45 hours of audio in approximately 800 hours of work, corresponding to a \textbf{17×} real-time factor (17 hours of annotation per hour of speech). Using \texttt{soloni-v2} instead of \texttt{soloni-v0} yielded a 112\% improvement in the rate at which human-corrected transcriptions could be produced, compared to an earlier study in which a 36× ratio was observed \cite{diarra2025costanalysishumancorrectedtranscription}.

\section{Conclusion}
\label{sec:conclusion}
We released 612 hours of spontaneous Bambara speech and a 423-hour segmented corpus, together with metadata, VAD-based SNR estimates, transcription guidelines, a minimalist mobile recording app, multiple evaluation sets, and monolingual ASR models (an ultra-compact QuartzNet variant and a 114M-parameter Parakeet-based model). For a low-literacy, predominantly oral language, this substantially increases the pool of publicly available resources.

Using \texttt{soloni-114m-tdt-ctc-v0} in a human–model loop enabled semi-automated annotation, reducing the time required for corrected transcriptions and improving WER/CER on both in-domain and more challenging benchmarks. Human evaluation on Nyana Eval revealed systematic gaps between automatic string-based metrics and native-speaker judgments.

Taken together, these elements define a practical workflow for dealing with the hard facts of low-resource African NLP: targeted field collection with trained facilitators, noise-aware preprocessing, semi-automated annotation, and evaluation protocols that combine automatic and human measures while accounting for deployment on modest hardware. Future work will extend this approach to other Manding languages and to datasets that more fully reflect real-world speech.

\section{Limitations}
\label{sec:limitations}
The African Next Voices (ANV) Bambara dataset was meticulously designed to provide a high-quality, clean foundation for ASR research. While successful in establishing the largest open corpus for the language, this intentional simplification inherently introduces limitations when considering the deployment of derived models in authentic, unconstrained Malian contexts. The core limitation lies in the necessary trade-off between controlled data collection and the complex nature of the real-world environment. We can enumerate the following key limitations:

\begin{itemize}
    \item \textbf{Acoustic Purity vs Real-World challenges:} The emphasis on clean recordings resulted in a dataset where 71.75\% of the recordings were classified with 'High SNR' or 'Very High SNR'. This limits the model’s exposure to challenging acoustic conditions—such as urban street noise and background voices that characterize typical Malian environments. Consequently, models trained exclusively on this corpus may exhibit a noticeable drop in performance when deployed in the real world.
    \item \textbf{Suppression of Code-Switching and Multilingualism:} The transcription protocol's mandate to replace code-switched terms and foreign words with the generic $\texttt{[cs]}$ tag, or to force a Bambara-phonology transcription, simplifies the target vocabulary for ASR training. However, this approach sacrifices linguistic realism. The resulting models are fundamentally unprepared to transcribe the common, fluid shifts between Bambara and French, which are integral to spontaneous speech.
\end{itemize}

Ultimately, the ANV Bambara dataset represents a simplified version of the language's acoustic and linguistic reality. While this simplification provides a more stable foundation for core ASR research, it comes at the cost of real-world robustness. Practitioners seeking to deploy these models in authentic Malian contexts characterized by inherent noise, fluent code-switching, and diverse accents must anticipate the need for a targeted domain adaptation.

\bibliography{custom}

@article{vad_snr,
author = {Vondrasek, Martin and Pollák, Petr},
year = {2005},
month = {04},
pages = {},
title = {Methods for Speech SNR estimation: Evaluation Tool and Analysis of VAD Dependency},
volume = {14},
journal = {Radioengineering}
}

@INPROCEEDINGS{cmuwilderness_black,
  author={Black, Alan W},
  booktitle={ICASSP 2019 - 2019 IEEE International Conference on Acoustics, Speech and Signal Processing (ICASSP)}, 
  title={CMU Wilderness Multilingual Speech Dataset}, 
  year={2019},
  volume={},
  number={},
  pages={5971-5975},
  doi={10.1109/ICASSP.2019.8683536}
}

@book{ethnologue2023,
  editor    = {Eberhard, David M. and Simons, Gary F. and Fennig, Charles D.},
  title     = {Ethnologue: Languages of the World},
  edition   = {26th},
  address   = {Dallas, TX},
  publisher = {SIL International},
  year      = {2023},
  url       = {https://www.ethnologue.com}
}

@misc{Marivate2025African,
  author       = {Marivate, Vukosi and Adebara, Ife and Wanzare, Lilian},
  title        = {African languages for AI: the project that's gathering a huge new dataset},
  howpublished = {The Conversation},
  url          = {https://theconversation.com/african-languages-for-ai-the-project-thats-gathering-a-huge-new-dataset-266371},
  month        = oct,
  year         = {2025},
  note         = {Accessed: 2025-11-13}
}

@misc{label2020studio,
  title={{Label Studio}: Data labeling software},
  url={https://github.com/heartexlabs/label-studio},
  note={Open source software available from https://github.com/heartexlabs/label-studio},
  author={
    Maxim Tkachenko and
    Mikhail Malyuk and
    Andrey Holmanyuk and
    Nikolai Liubimov},
  year={2020-2022},
}

@misc{emezue2025naijavoicesdatasetcultivatinglargescale,
      title={The NaijaVoices Dataset: Cultivating Large-Scale, High-Quality, Culturally-Rich Speech Data for African Languages}, 
      author={Chris Emezue and NaijaVoices Community and Busayo Awobade and Abraham Owodunni and Handel Emezue and Gloria Monica Tobechukwu Emezue and Nefertiti Nneoma Emezue and Sewade Ogun and Bunmi Akinremi and David Ifeoluwa Adelani and Chris Pal},
      year={2025},
      eprint={2505.20564},
      archivePrefix={arXiv},
      primaryClass={cs.CL},
      url={https://arxiv.org/abs/2505.20564},
}

@inproceedings{ardila-etal-2020-common,
    title = "{Common Voice}: A Massively-Multilingual Speech Corpus",
    author = "Ardila, Rosana  and
      Branson, Megan  and
      Davis, Kelly  and
      Kohler, Michael  and
      Meyer, Josh  and
      Henretty, Michael  and
      Morais, Reuben  and
      Saunders, Lindsay  and
      Tyers, Francis M.  and
      Weber, Gregor",
    editor = "Calzolari, Nicoletta  and
      Sutcliffe, Richard  and
      Hajic, Jan  and
      Maire, Fr{\'e}d{\'e}ric",
    booktitle = "Proceedings of The 12th Language Resources and Evaluation Conference (LREC 2020)",
    month = may,
    year = "2020",
    address = "Marseille, France",
    publisher = "European Language Resources Association (ELRA)",
    url = "https://aclanthology.org/2020.lrec-1.520",
    pages = "4211--4215"
}

@misc{Diarra2022Griots,
  author                = {Sebastien Diarra and Michael Leventhal and Allahsera Auguste Tapo},
  title                 = {RobotsMali Griots Speech Dataset, and ASR},
  howpublished          = {\url{https://github.com/robotsmali-ai/jeli-asr/}},
  year                  = {2022},
}

@misc{omnilingualasr2025,
    title={{Omnilingual ASR}: Open-Source Multilingual Speech Recognition for 1600+ Languages},
    author={{Omnilingual ASR Team} and Keren, Gil and Kozhevnikov, Artyom and Meng, Yen and Ropers, Christophe and Setzler, Matthew and Wang, Skyler and Adebara, Ife and Auli, Michael and Chan, Kevin and Cheng, Chierh and Chuang, Joe and Droof, Caley and Duppenthaler, Mark and Duquenne, Paul-Ambroise and Erben, Alexander and Gao, Cynthia and Mejia Gonzalez, Gabriel and Lyu, Kehan and Miglani, Sagar and Pratap, Vineel and Sadagopan, Kaushik Ram and Saleem, Safiyyah and Turkatenko, Arina and Ventayol-Boada, Albert and Yong, Zheng-Xin and Chung, Yu-An and Maillard, Jean and Moritz, Rashel and Mourachko, Alexandre and Williamson, Mary and Yates, Shireen},
    year={2025},
    url={https://ai.meta.com/research/publications/omnilingual-asr-open-source-multilingual-speech-recognition-for-1600-languages/},
}

@article{goyal-etal-2022-flores,
    title = "The {F}lores-101 Evaluation Benchmark for Low-Resource and Multilingual Machine Translation",
    author = "Goyal, Naman  and
      Gao, Cynthia  and
      Chaudhary, Vishrav  and
      Chen, Peng-Jen  and
      Wenzek, Guillaume  and
      Ju, Da  and
      Krishnan, Sanjana  and
      Ranzato, Marc{'}Aurelio  and
      Guzm{\'a}n, Francisco  and
      Fan, Angela",
    editor = "Roark, Brian  and
      Nenkova, Ani",
    journal = "Transactions of the Association for Computational Linguistics",
    volume = "10",
    year = "2022",
    address = "Cambridge, MA",
    publisher = "MIT Press",
    url = "https://aclanthology.org/2022.tacl-1.30/",
    doi = "10.1162/tacl_a_00474",
    pages = "522--538"
}

@misc{pratap2023scalingspeechtechnology1000,
      title={Scaling Speech Technology to 1,000+ Languages}, 
      author={Vineel Pratap and Andros Tjandra and Bowen Shi and Paden Tomasello and Arun Babu and Sayani Kundu and Ali Elkahky and Zhaoheng Ni and Apoorv Vyas and Maryam Fazel-Zarandi and Alexei Baevski and Yossi Adi and Xiaohui Zhang and Wei-Ning Hsu and Alexis Conneau and Michael Auli},
      year={2023},
      eprint={2305.13516},
      archivePrefix={arXiv},
      primaryClass={cs.CL},
      url={https://arxiv.org/abs/2305.13516}, 
}

@misc{baevski2020wav2vec20frameworkselfsupervised,
      title={wav2vec 2.0: A Framework for Self-Supervised Learning of Speech Representations}, 
      author={Alexei Baevski and Henry Zhou and Abdelrahman Mohamed and Michael Auli},
      year={2020},
      eprint={2006.11477},
      archivePrefix={arXiv},
      primaryClass={cs.CL},
      url={https://arxiv.org/abs/2006.11477}, 
}

@article{humanparityASR,
author = {Xiong, Wayne and Droppo, Jasha and Huang, Xuedong and Seide, Frank and Seltzer, Michael and Stolcke, Andreas and Yu, Dong and Zweig, Geoffrey},
year = {2016},
month = {10},
pages = {},
title = {Achieving Human Parity in Conversational Speech Recognition},
volume = {PP},
journal = {IEEE/ACM Transactions on Audio, Speech, and Language Processing},
doi = {10.1109/TASLP.2017.2756440}
}

@misc{kaplan2020scalinglawsneurallanguage,
      title={Scaling Laws for Neural Language Models}, 
      author={Jared Kaplan and Sam McCandlish and Tom Henighan and Tom B. Brown and Benjamin Chess and Rewon Child and Scott Gray and Alec Radford and Jeffrey Wu and Dario Amodei},
      year={2020},
      eprint={2001.08361},
      archivePrefix={arXiv},
      primaryClass={cs.LG},
      url={https://arxiv.org/abs/2001.08361}, 
}

@misc{SileroVAD,
  author = {{Silero Team}},
  title = {Silero VAD: pre-trained enterprise-grade Voice Activity Detector (VAD), Number Detector and Language Classifier},
  year = {2024},
  publisher = {GitHub},
  howpublished = {\url{https://github.com/snakers4/silero-vad}},
  email = {hello@silero.ai}
}

@misc{li2019santlrspeechannotationtoolkit,
      title={SANTLR: Speech Annotation Toolkit for Low Resource Languages}, 
      author={Xinjian Li and Zhong Zhou and Siddharth Dalmia and Alan W. Black and Florian Metze},
      year={2019},
      eprint={1908.01067},
      archivePrefix={arXiv},
      primaryClass={cs.CL},
      url={https://arxiv.org/abs/1908.01067}, 
}

@inproceedings{xu05b_interspeech,
  title     = {Robust speech recognition based on noise and SNR classification - a multiple-model framework},
  author    = {Haitian Xu and Zheng-Hua Tan and Paul Dalsgaard and Børge Lindberg},
  year      = {2005},
  booktitle = {Interspeech 2005},
  pages     = {977--980},
  doi       = {10.21437/Interspeech.2005-233},
  issn      = {2958-1796},
}

@inproceedings{tapo-etal-2025-gaife,
    title = "{GAI}f{E}: Using {G}en{AI} to Improve Literacy in Low-resourced Settings",
    author = "Tapo, Allahsera Auguste  and
      Coulibaly, Nouhoum  and
      Diallo, Seydou  and
      Diarra, Sebastien  and
      Homan, Christopher M  and
      Keita, Mamadou K.  and
      Leventhal, Michael",
    editor = "Chiruzzo, Luis  and
      Ritter, Alan  and
      Wang, Lu",
    booktitle = "Findings of the Association for Computational Linguistics: NAACL 2025",
    month = apr,
    year = "2025",
    address = "Albuquerque, New Mexico",
    publisher = "Association for Computational Linguistics",
    url = "https://aclanthology.org/2025.findings-naacl.442/",
    doi = "10.18653/v1/2025.findings-naacl.442",
    pages = "7914--7929",
    ISBN = "979-8-89176-195-7"
}

@misc{diarra_kunkado_2025,
  title        = {kunnafonidilaw ka cadeau: an {ASR} dataset to power the development of models that understand present-Day Bambara},
  author       = {Diarra, Yacouba and Coulibaly, Nouhoum and Kamaté, Panga Azazia and Leventhal, Michael},
  year         = 2025,
  howpublished = {Hugging Face Datasets},
  note         = {Arxiv coming soon},
  url          = {https://huggingface.co/datasets/RobotsMali/kunkado}
}

@misc{kriman2019quartznetdeepautomaticspeech,
      title={QuartzNet: Deep Automatic Speech Recognition with 1D Time-Channel Separable Convolutions}, 
      author={Samuel Kriman and Stanislav Beliaev and Boris Ginsburg and Jocelyn Huang and Oleksii Kuchaiev and Vitaly Lavrukhin and Ryan Leary and Jason Li and Yang Zhang},
      year={2019},
      eprint={1910.10261},
      archivePrefix={arXiv},
      primaryClass={eess.AS},
      url={https://arxiv.org/abs/1910.10261}, 
}

@misc{xu2023efficientsequencetransductionjointly,
      title={Efficient Sequence Transduction by Jointly Predicting Tokens and Durations}, 
      author={Hainan Xu and Fei Jia and Somshubra Majumdar and He Huang and Shinji Watanabe and Boris Ginsburg},
      year={2023},
      eprint={2304.06795},
      archivePrefix={arXiv},
      primaryClass={eess.AS},
      url={https://arxiv.org/abs/2304.06795}, 
}

@misc{rekesh2023fastconformerlinearlyscalable,
      title={Fast Conformer with Linearly Scalable Attention for Efficient Speech Recognition}, 
      author={Dima Rekesh and Nithin Rao Koluguri and Samuel Kriman and Somshubra Majumdar and Vahid Noroozi and He Huang and Oleksii Hrinchuk and Krishna Puvvada and Ankur Kumar and Jagadeesh Balam and Boris Ginsburg},
      year={2023},
      eprint={2305.05084},
      archivePrefix={arXiv},
      primaryClass={eess.AS},
      url={https://arxiv.org/abs/2305.05084}, 
}

@misc{zupon2021textnormalizationlowresourcelanguages,
      title={Text Normalization for Low-Resource Languages of Africa}, 
      author={Andrew Zupon and Evan Crew and Sandy Ritchie},
      year={2021},
      eprint={2103.15845},
      archivePrefix={arXiv},
      primaryClass={cs.CL},
      url={https://arxiv.org/abs/2103.15845}, 
}

@misc{vaswani2023attentionneed,
      title={Attention Is All You Need}, 
      author={Ashish Vaswani and Noam Shazeer and Niki Parmar and Jakob Uszkoreit and Llion Jones and Aidan N. Gomez and Lukasz Kaiser and Illia Polosukhin},
      year={2017},
      eprint={1706.03762},
      archivePrefix={arXiv},
      primaryClass={cs.CL},
      url={https://arxiv.org/abs/1706.03762}, 
}

@misc{ginsburg2020stochasticgradientmethodslayerwise,
      title={Stochastic Gradient Methods with Layer-wise Adaptive Moments for Training of Deep Networks}, 
      author={Boris Ginsburg and Patrice Castonguay and Oleksii Hrinchuk and Oleksii Kuchaiev and Vitaly Lavrukhin and Ryan Leary and Jason Li and Huyen Nguyen and Yang Zhang and Jonathan M. Cohen},
      year={2020},
      eprint={1905.11286},
      archivePrefix={arXiv},
      primaryClass={cs.LG},
      url={https://arxiv.org/abs/1905.11286}, 
}

@misc{graves2012sequencetransductionrecurrentneural,
      title={Sequence Transduction with Recurrent Neural Networks}, 
      author={Alex Graves},
      year={2012},
      eprint={1211.3711},
      archivePrefix={arXiv},
      primaryClass={cs.NE},
      url={https://arxiv.org/abs/1211.3711}, 
}

@misc{li2020comparisonpopularendtoendmodels,
      title={On the Comparison of Popular End-to-End Models for Large Scale Speech Recognition}, 
      author={Jinyu Li and Yu Wu and Yashesh Gaur and Chengyi Wang and Rui Zhao and Shujie Liu},
      year={2020},
      eprint={2005.14327},
      archivePrefix={arXiv},
      primaryClass={eess.AS},
      url={https://arxiv.org/abs/2005.14327}, 
}

@inproceedings{gravesctc2006,
author = {Graves, Alex and Fernández, Santiago and Gomez, Faustino and Schmidhuber, Jürgen},
year = {2006},
month = {01},
pages = {369-376},
title = {Connectionist temporal classification: Labelling unsegmented sequence data with recurrent neural 'networks},
volume = {2006},
booktitle = {ICML 2006},
journal = {ICML 2006 - Proceedings of the 23rd International Conference on Machine Learning},
doi = {10.1145/1143844.1143891}
}

@article{konta2014,
  author    = {Konta, Mamadou and Vydrin, Valentin},
  title     = {Propositions pour l’orthographe du bamanankan},
  journal   = {Mandenkan},
  number    = {52},
  pages     = {3--38},
  year      = {2014}
}

@article{vydrin:halshs-03909864,
  TITLE = {{Vers un dictionnaire orthographique bambara}},
  AUTHOR = {Vydrin, Valentin Feodosievich},
  URL = {https://shs.hal.science/halshs-03909864},
  JOURNAL = {{Mandenkan : Bulletin Semestriel d'{\'E}tudes Linguistiques Mand{\'e}}},
  PUBLISHER = {{Presses de l'Inalco}},
  NUMBER = {68},
  PAGES = {59-82},
  YEAR = {2022},
  MONTH = Dec,
  DOI = {10.4000/mandenkan.2905},
  HAL_ID = {halshs-03909864},
  HAL_VERSION = {v1},
}

@misc{diarra2025costanalysishumancorrectedtranscription,
      title={Cost Analysis of Human-corrected Transcription for Predominately Oral Languages}, 
      author={Yacouba Diarra and Nouhoum Souleymane Coulibaly and Michael Leventhal},
      year={2025},
      eprint={2510.12781},
      archivePrefix={arXiv},
      primaryClass={cs.CL},
      url={https://arxiv.org/abs/2510.12781}, 
}

@misc{tall_2025_17672774,
author = {Tall, Madani Amadou},
title = {Analyse comparative humaine des modèles ASR Bambara de RobotsMali},
month = nov,
year = 2025,
publisher = {Zenodo},
doi = {10.5281/zenodo.17672774},
url = {https://doi.org/10.5281/zenodo.17672774},
}

\appendix
\section{Transcription Guidelines}
\label{sec:trans-guide}

\begin{figure*}[t]
    \centering
    \includegraphics[width=\linewidth]{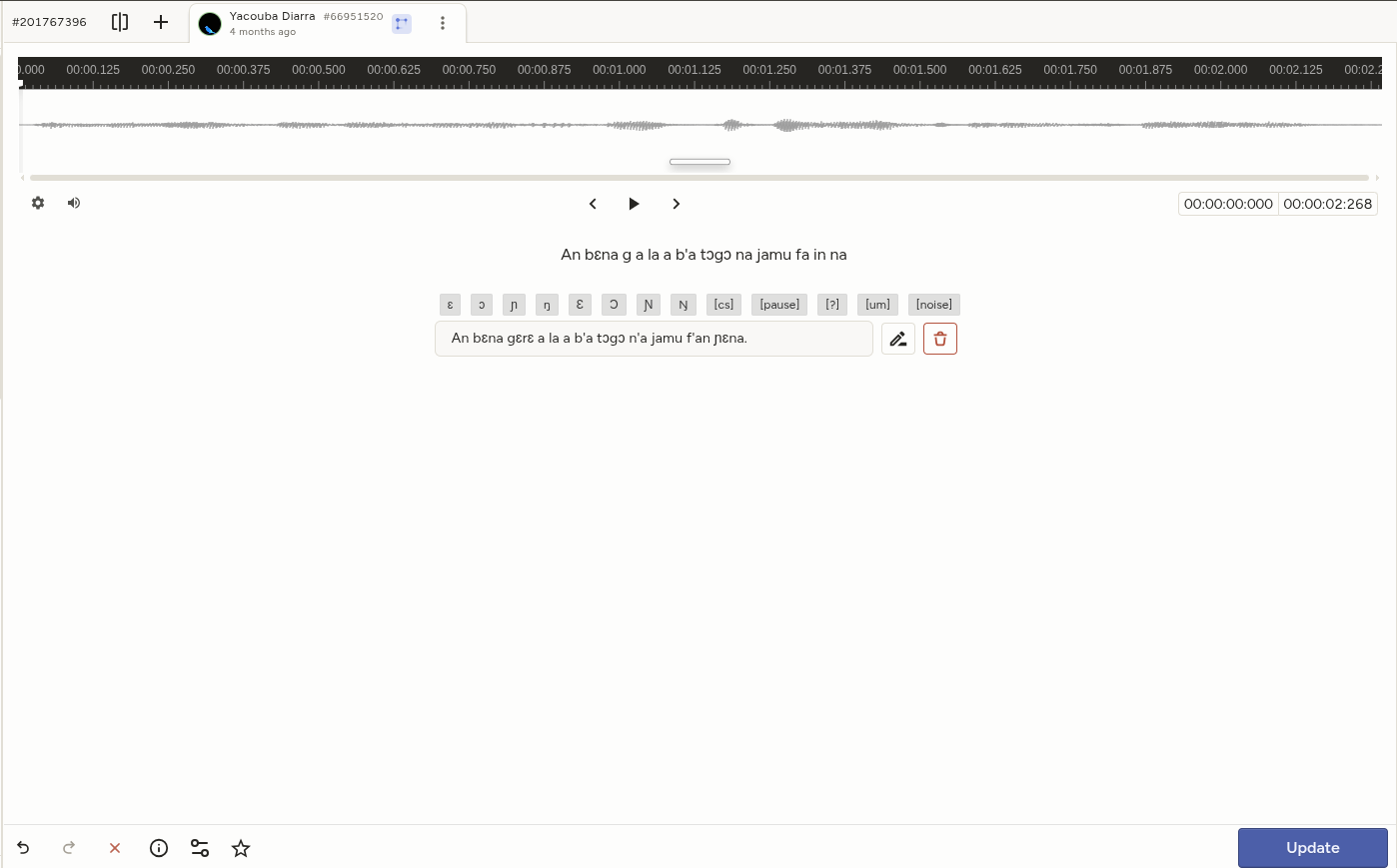}
    \caption{The Labeling Interface for the African Next Voices Bambara Transcription Project. The interface shows the original audio waveform, the automatically generated pre-transcription, and the field for human correction/validation.}
    \label{fig:labeling-interface}
\end{figure*}

The transcription process for this Bambara dataset was implemented as a \textbf{First Review} task, where annotators corrected and validated pre-transcribed audio segments. Annotators were instructed to respect the following rules to simplify and standardize transcription and ensure high-quality data for ASR training and evaluation.

\subsection*{Language and Orthography}
\begin{itemize}
    \item Use the \textit{standardized orthography of Bambara} \cite{konta2014}. Correct any orthographic errors or non-standard characters (e.g., accented characters are prohibited, conforming to the ordinance of AMALAN\footnote{We have been closing working with the Académie Malienne des Langues (AMALAN) and the Direction Nationale de l'Education Non Formelle et des Langues Nationales (DNENF-LN) prior to this project and we were able to leverage their expertise for all matters pertaining to national language standards for Bambara.}).
    \item Do not correct the speaker's grammatical errors or pronunciation mistakes.
    \item Elision and Mispronunciation: Transcribe exactly what is pronounced. For example, transcribe "Ne b'a  \textipa{f\textepsilon}" instead of "Ne be a \textipa{f\textepsilon}" if "b'a" was pronounced.
    \item Proper Nouns: Must be capitalized.
    \item Repetitions, Disfluencies, and Stuttering: Write out repeated words/phonemes without using ellipses (e.g., "n'i ko ko ko ne kalabanci e dun?").
    \item Consult the Bamadaba dictionary for language references \cite{vydrin:halshs-03909864}.
\end{itemize}

\subsection*{Numbers, Abbreviations, and Foreign Words}
\begin{itemize}
    \item Numbers: Must be written in full letters exactly as pronounced (e.g., 35 $\to$ *bi saba ni duuru / bi saba ni Loru*).
    \item The Ordinal forms (e.g., *folo*, *filanan*, *sabanan*) must also be written in full letters.
    \item If a number is pronounced in a foreign language (e.g., French), use the code-switching tag $\texttt{[cs]}$.
    \item Abbreviations and Acronyms: Transcribe them as pronounced and in \textit{uppercase} (e.g., FIFA, BIM, ORTM). Do not use periods in acronyms.
    \item Spelled Words: Use hyphens to separate letters (e.g., $\texttt{K-E-L-E-N}$).
    \item \textbf{Code-Switching/Foreign Words:}
    \begin{itemize}
        \item If the word exists in Bamadaba and was pronounced with the local Bambara phonology, transcribe it as written in the dictionary (e.g., Parce que $\to$ *paseke*, passerelle $\to$ *pasereli*). This rule exists because Bambara, like many African languages, has borrowed, transformed and standardized many words from the former colonial language.
        \item If the word was pronounced with its original foreign pronunciation and/or is not recognized in Bamadaba, replace it with the tag $\texttt{[cs]}$ (e.g., $\texttt{ne ka véhicule} \to \texttt{ne ka [cs]}$).
    \end{itemize}
\end{itemize}

\subsection*{Acoustic Event Tags (for ASR Modeling)}
Acoustic events and background sounds were retained and modeled in the final transcriptions using the following specific tags:

\begin{itemize}
    \item Inaudible/Incomplete Speech or Overlap: Use the tag $\texttt{[?]}$ for incomprehensible, non-audible speech, or speech overlaps.
    \item Vocalized/Disfluencies (Mouth Sounds): Use the tag $\texttt{[um]}$ for sounds like *\textipa{\textepsilon\textepsilon}*, *hum*, *onh*, *ah*, *unhun*, etc..
    \item Long Silences: Use the tag $\texttt{[pause]}$ for silences longer than 5 seconds (or longer than 3 seconds at the beginning or end of a segment). This tag was rarely used after VAD segmentation. 
    \item Background Noise: Use the tag $\texttt{[noise]}$ for all occurrence strong background noise, including applause, coughing, laughter, phone rings, children, etc..
\end{itemize}

\subsection*{Punctuation}
\begin{itemize}
    \item Standard punctuation (commas, periods, question marks, etc.) should be used.
\end{itemize}

\section{Labeling Interface}
\label{sec:labeling}
The data annotation for the ANV Bambara project was performed using a tailored platform built on top of Label Studio \cite{label2020studio}. This interface facilitated the task by presenting pre-transcribed audio segments for human correction and validation. The audio files and their pre-labels were sequentially loaded from Google Cloud Storage into the labeling interface.

As shown in Figure \ref{fig:labeling-interface}, the simple interface provided the following key elements:

\begin{itemize}
    \item Audio Segment: The interface displays the audio waveform and playback controls, allowing the annotator to listen to the segment.
    \item Pre-Transcription: The initial transcription was automatically generated by our ASR models.
    \item Correction Field: The annotator validates and corrects the automatic transcription in a designated field.
    \item Acoustic Event Tags: A row of buttons provides quick access to the acoustic event tags and the few Bambara characters that are not typically found on a standard keyboard.
\end{itemize}

This process of pre-transcription followed by human correction minimized human labor for segmentation while optimizing the usable speech data for the transcription pipeline.
\end{document}